# A Comparative Benchmark of a Moroccan Darija Toxicity Detection Model (Typica.ai) and Major LLM-Based Moderation APIs (OpenAI, Mistral, Anthropic)

Hicham Assoudi[1],[2]
Typica.ai[3]
Montreal, Canada
assoudi@typica.ai

## ABSTRACT
This paper presents a comparative benchmark evaluating the performance of Typica.ai's custom Moroccan Darija toxicity detection model against major LLM-based moderation APIs: OpenAI (omni-moderation-latest), Mistral (mistral-moderation-latest), and Anthropic Claude (claude-3-haiku-20240307). We focus on culturally grounded toxic content, including implicit insults, sarcasm, and culturally specific aggression often overlooked by general-purpose systems. Using a balanced test set derived from the OMCD_Typica.ai_Mix dataset, we report precision, recall, F1-score, and accuracy, offering insights into challenges and opportunities for moderation in underrepresented languages. Our results highlight Typica.ai's superior performance, underlining the importance of culturally adapted models for reliable content moderation.

## 1. INTRODUCTION
With the proliferation of digital platforms, the need for reliable and culturally sensitive content moderation systems has escalated dramatically. While many moderation approaches today utilize multilingual large language models (LLM), they are typically optimized for high-resource languages and often lack the capacity to understand region-specific nuances. These generic models frequently struggle when confronted with culturally specific, informal dialects and context-dependent expressions. Moroccan Darija, a dialect spoken daily by millions, exemplifies this challenge with its heavy code-switching, fluid orthography, and deeply embedded cultural idioms and insults that defy literal translation. Addressing these gaps is essential not only for technical performance but also for ethical AI deployment that respects local linguistic identities. The current research thus presents the first rigorous and systematic evaluation comparing commercially available moderation APIs—specifically those by OpenAI, Mistral , and Anthropic Claude—against a a culturally and domain-specific toxicity model developed by Typica.ai, explicitly designed to capture the nuances and sensitivities of Moroccan Darija.

## 2. RELATED WORK
Existing research highlights the frequent shortcomings of multilingual language models (LLMs) when applied across diverse linguistic and cultural contexts. Recent studies show that although ChatGPT achieves excellent results on English NLP tasks, it is consistently outperformed by smaller models fine-tuned specifically on Arabic, across a wide range of benchmarks [1]. Faisal et al. [2] further demonstrate that while LLMs show sensitivity to multilingual and dialectal variations in toxicity detection, their weakest area remains consistency with human judgments — exposing a critical limitation in relying solely on LLM-as-a-judge frameworks for dialectal evaluation.

Importantly, Arabic dialect processing presents unique challenges due to linguistic variability, lack of standardized resources, and cultural nuances that general-purpose models often fail to capture. While multilingual or Modern Standard Arabic (MSA)-focused models perform reasonably, their effectiveness drops significantly when handling dialectal varieties like Moroccan Darija. This has led to several important efforts to develop dialect-specific datasets and models.

For Moroccan Darija, Rachidi et al. [3] developed machine and deep learning models to classify toxicity on Instagram, but their results highlight limitations driven by the scarcity of high-quality datasets and preprocessing tools. Zarnoufi et al. [4] introduced MAOffens, the first balanced Moroccan Arabic offensive language dataset, enabling transformer-based models to better detect social media offensiveness. Ibrahimi and Mourhir [5] contributed a human-labeled dataset of over 20,000 Darija sentences for offensive language detection, providing an essential labeled resource for the research community. Essefar et al. [6] introduced OMCD, a large annotated Moroccan offensive comments dataset, allowing for systematic evaluation of machine learning and deep learning models on Moroccan dialect.

While these foundational works have laid critical groundwork, they predominantly focus on model development and dataset curation. A key gap remains in comparative evaluations between these dialect-specific resources and commercial moderation systems (e.g., OpenAI, Mistral, Anthropic Claude). Without such comparative benchmarks, it is difficult to assess how well widely adopted moderation APIs perform on culturally complex, low-resource languages like Moroccan Darija — leaving open questions about fairness, cultural alignment, and real-world robustness.

---

[1] Founder at Typica.ai.

[2] External Research Associate at CRIA (Centre de Recherche en Intelligence Artificielle), Université du Québec à Montréal (UQAM).

[3] Typica.ai is an AI research startup specializing in culturally localized AI solutions

## 3. DATASET
This section presents an overview of the evaluation dataset, detailing its composition and highlighting its critical role in validating the comparative experiments. We explain why the OMCD dataset was selected, how it was enriched, and what criteria guided the validation process — all to strengthen the transparency and scientific rigor of our evaluation.

### 3.1 Composition
To enable a rigorous evaluation, we compiled a carefully constructed dataset combining public and proprietary Moroccan Darija examples. This dataset was built upon the well-known OMCD (Offensive Moroccan Comments Dataset) [6], widely recognized as a foundational resource for toxicity detection research in Darija due to its size, diversity, and solid annotation protocols. We selected OMCD as the backbone for validation because it provides a reliable, widely cited benchmark that ensures our evaluation aligns with prior work. However, we identified limitations in its coverage of more nuanced, culturally embedded toxic expressions, particularly indirect insults, sarcasm, and euphemisms that general-purpose models frequently overlook. To address this, we created an enriched version, OMCD_Typica.ai_Mix, by integrating Typica.ai's proprietary annotations and expanding the label space to better reflect the cultural and linguistic subtleties of Moroccan online discourse. This enriched dataset serves as a more comprehensive and challenging benchmark, significantly enhancing the robustness and fairness of our comparative evaluation between general-purpose moderation APIs and our specialized Typica.ai classifier.

### 3.2 Preprocessing
Before evaluation, we applied a comprehensive filtering and cleaning pipeline to ensure that input sentences were representative of Moroccan Darija and free from confounding artifacts such as Latin script noise or metadata tokens. Specifically, we removed any sentence where more than 50% of the characters were in Latin script, aiming to avoid misclassifying code-switched examples dominated by French or English, which are outside the scope of our Darija-focused classifier. Additionally, we applied a cleaning function that removed Latin characters, symbols, and punctuation. This preprocessing step was essential to standardize Darija text inputs while preserving their linguistic meaning, ensuring that the final dataset, OMCD_Typica.ai_Mix, retained both linguistic diversity and cultural relevance, providing a strong foundation for building robust, culturally aligned toxicity classifiers for Moroccan Darija.

### 3.3 Balancing
To address class imbalance and prevent model bias toward the majority class (non-toxic), we applied random undersampling to the dataset. Specifically, we preserved all examples labeled as toxic (label=1) and randomly sampled an equal number of non-toxic (label=0) examples, ensuring a 1:1 ratio between toxic and clean examples. After balancing and cleaning, we split the resulting dataset into training, validation, and test sets, with each subset reflecting the cleaned and balanced characteristics. Importantly, for this paper's benchmark and evaluation, we used only the test split of the OMCD_Typica.ai_Mix dataset as the definitive reference point for comparing the Typica.ai classifier against the OpenAI and Mistral moderation APIs. This focused comparative evaluation ensures consistency, fairness, and methodological rigor throughout the benchmarking phase.

In summary, the OMCD_Typica.ai_Mix dataset (12,758 Moroccan Darija comments) brings together public, curated, and proprietary sources, ensuring broad linguistic and cultural coverage. Only the test split was used for the benchmark evaluation of Typica.ai, OpenAI, Mistral, and Anthropic models; the train and validation splits were prepared but not used in this study. The dataset splits are summarized in Table 1.

Table 1. Dataset Splits

| Subset | Size | Purpose |
| --- | --- | --- |
| Train | 9,568 | Prepared but not used in this benchmark |
| Validation | 2,552 | Prepared but not used in this benchmark |
| **Test** | **638** | **Used for final benchmark evaluation** |

## 4. METHODOLOGY
To ensure fair, transparent, and reproducible evaluation, we designed a consistent methodology aligned with benchmark best practices. Our goal was to assess the relative strengths and weaknesses of three prominent commercial moderation APIs (OpenAI, Mistral, Anthropic Claude) and Typica.ai's specialized Moroccan Darija toxicity model, with particular focus on detecting culturally specific expressions often overlooked by general-purpose systems.

All models and APIs were evaluated on the same set of test inputs from the OMCD_Typica.ai_Mix test split. Outputs were analyzed using a unified evaluation script computing standard metrics — precision, recall, F1-score, and accuracy — focused specifically on the toxic (flagged) class [7]. This approach ensured consistent result interpretation and minimized biases from differences in input formatting, preprocessing, or postprocessing.

Beyond overall performance, we critically examined how each system handled domain-specific and culturally embedded toxicity patterns, combining quantitative metrics with qualitative insights for a robust comparative perspective.

### 4.1 OpenAI Moderation API
We tested the OpenAI Moderation API using the latest moderation model (omni-moderation-latest), which leverages OpenAI's most recent multimodal GPT-4o moderation model capable of handling both text and image inputs [8], [9]. According to OpenAI's official documentation, this new model improves accuracy across multiple languages, adding more granular harm categories and better probability calibration compared to earlier versions[8]. It detects categories such as hate, hate/threatening, harassment, harassment/threatening, sexual, sexual/minors, violence, violence/graphic, illicit, illicit/violent, and self-harm (including intent and instructions). The API returns a structured JSON object containing a global flagged field and per-category binary flags along with confidence scores. Although highly effective for English and several other supported languages, the model has not been explicitly tested or adapted for Moroccan Darija. For this evaluation, we treated the overall flagged field as the primary moderation decision when comparing outputs across systems.

### 4.2 Mistral Moderation API
We used the Mistral client to call the Mistral moderation API, specifically invoking the Mistral moderation model (mistral-moderation-latest) through the client interface [10], [11]. According to Mistral's official description, this is the same moderation system that powers the moderation layer in their Le Chat product, designed to provide system-level guardrails for downstream deployments and empower users to tailor moderation to specific applications and safety standards [10]. The API returns category scores across several dimensions, including sexual, hate and discrimination, violence and threats, dangerous and criminal content, self-harm, health, financial, law, and personally identifiable information (PII). For evaluation, we inferred toxicity if any of the main categories (e.g., violence, hate, harassment) produced a positive (true) flag. We simplified the evaluation logic by treating the result as toxic (flagged) if any category (such as violence, hate, or harassment) returned a true flag, ensuring consistency and alignment across systems.

### 4.3 Anthropic Claude Moderation Model

We included Anthropic Claude (claude-3-haiku-20240307), a prompt-driven LLM moderation system following the approach outlined in Anthropic's Content Moderation Cookbook [12], [13]. Unlike dedicated moderation APIs, Claude relies on detailed prompts where moderation rules and categories (e.g., BLOCK and ALLOW) are explicitly defined by the user, making it highly flexible but also introducing variability depending on prompt design. In our setup, we explicitly used the basic prompt structure outlined in Anthropic's Content Moderation Cookbook [13], ensuring no additional adjustments were introduced that might bias results to ensure fair alignment with binary toxic/not-toxic outcomes, allowing a meaningful comparison against other systems. This section emphasizes that Claude's results hinge on prompt engineering rather than a fixed, pretrained moderation endpoint.

### 4.4 Typica.ai Darija Toxicity Classifier

Typica.ai developed a custom Moroccan Darija toxicity classifier specifically tailored to local linguistic and cultural nuances. The model is based on a fine-tuned BERT-based transformer and was trained on a manually curated internal dataset assembled by Typica.ai. Designed as a multi-class classifier, it predicts categories such as clean, hate, insult, offensive, obscene, among others.

For the purposes of this evaluation, we simplified the output by treating any predicted category other than clean as a flagged (toxic) result. This binarization ensured that the evaluation logic aligned with how the outputs of the evaluated commercial APIs were interpreted, enabling fair and consistent comparison across all systems.

### 4.5 Language Support

This subsection summarizes the declared language support of the evaluated systems. OpenAI's moderation model (omni-moderation-latest) reports improved multilingual performance, with up to 6× gains across 40 languages compared to earlier models, particularly in low-resource languages such as Khmer or Swati [8]; however, Moroccan Darija remains explicitly unsupported and untested. Mistral's moderation model (mistral-moderation-latest) is a natively multilingual LLM-based classifier, explicitly trained on languages including Arabic, French, Spanish, German, and others [10], yet it lacks fine-tuning for Moroccan Darija specifically.

Anthropic Claude (claude-3-haiku-20240307) offers a flexible, prompt-driven generic model that can be configured for moderation tasks and, according to Anthropic, demonstrates robust multilingual performance, including Arabic, while maintaining strong cross-lingual accuracy relative to English [14]. Notably, Claude's evaluations highlight particularly strong zero-shot performance across both widely spoken and lower-resource languages.

In contrast, Typica.ai's model is explicitly trained on Moroccan Darija (in Arabic script), which may offer an advantage in culturally aligned toxicity detection compared to general-purpose multilingual models or APIs.

### 4.6 Evaluation Metrics

We employed standard binary classification metrics — precision, recall, F1-score, and overall accuracy — calculated specifically for the toxic (flagged) class, which is the most critical category for our benchmark. Following common practices in moderation evaluation, all non-clean categories were aggregated into a single toxic class to focus on the system's ability to detect harmful content, regardless of subtype.

We report precision and recall to capture the essential trade-off between minimizing false positives (avoiding over-flagging toxic content) and minimizing false negatives (ensuring harmful content is caught). The F1-score offers a balanced summary of performance, particularly valuable under class imbalance, while overall accuracy is included for completeness, despite its known limitations in skewed datasets.

## 5. RESULTS

This section introduces the core evaluation results comparing the Typica.ai toxicity classifier against the OpenAI, Mistral, and Anthropic moderation APIs. We present both quantitative metrics and key performance insights that illustrate how each system performs on Moroccan Darija toxicity detection. Specifically, the evaluation included detailed precision, recall, and F1-scores for each system across both toxic and non-toxic classes. The final merged evaluation results covered 630 test examples, showing per-class precision, recall, F1-score, and macro and weighted averages, providing a robust, multi-dimensional view of comparative performance.

**Table 2. Class-wise Metrics (Accuracy, Macro F1, Toxic F1, Not Toxic F1)**

| Model | Accuracy | Macro F1 | Toxic F1 (True) | Not Toxic F1 (False) |
|---|---|---|---|---|
| Typica.ai | **0.830** | **0.830** | **0.834** | **0.827** |
| OpenAI | 0.652 | 0.644 | 0.589 | 0.699 |
| Mistral | 0.649 | 0.641 | 0.588 | 0.694 |
| Anthropic (Claude) | 0.659 | 0.617 | 0.743 | 0.492 |

Table 1 presents the class-wise performance metrics across all models, including overall accuracy, macro-averaged F1, and the per-class F1 scores for the toxic and non-toxic categories. Typica.ai achieves the highest accuracy (0.830) and macro F1 (0.830), with balanced performance across both toxic (0.834) and non-toxic (0.827) classes. In contrast, OpenAI and Mistral show notably lower accuracy (0.652 and 0.649, respectively) and struggle particularly on the toxic class, with F1 scores below 0.590. Anthropic Claude stands out for its high toxic F1 score (0.743), indicating strong detection of toxic content, but it suffers from very low non-toxic F1 (0.492), reflecting a tendency to over-flag clean content.

**Table 3. Overall Weighted Metrics (Precision, Recall, F1-score)**

| Model | Precision | Recall | F1-score |
|---|---|---|---|
| Typica.ai | **0.832** | **0.830** | **0.830** |
| OpenAI | 0.688 | 0.652 | 0.641 |
| Mistral | 0.682 | 0.649 | 0.639 |
| Anthropic (Claude) | 0.724 | 0.659 | 0.623 |

Table 2 summarizes the overall weighted precision, recall, and F1-score across all classes, providing an at-a-glance benchmark comparison. Typica.ai again leads with weighted precision (0.832), recall (0.830), and F1-score (0.830), outperforming all other models. Anthropic Claude, while showing slightly better weighted precision (0.724) than OpenAI (0.688) and Mistral (0.682), achieves the lowest weighted F1-score (0.623) due to its imbalance between precision and recall. OpenAI and Mistral show similar performance profiles, slightly trailing each other but both remaining well below Typica.ai's benchmark. These overall metrics confirm that Typica.ai consistently delivers the most robust and reliable performance across diverse moderation scenarios.

To further illustrate these findings, we present a comparative bar graph summarizing the weighted F1 score across all four models, providing a clear visual comparison that highlights Typica.ai's superior performance.

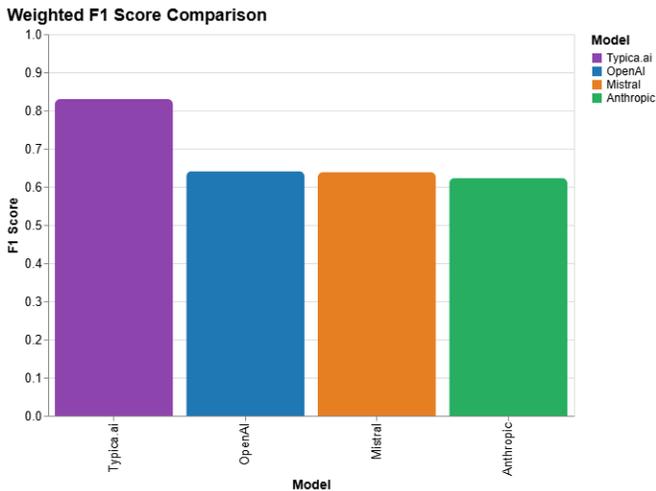

**Figure 1. Weighted F1 Score Comparison Across Typica.ai, OpenAI, Mistral, and Claude Models.**

To summarize, Typica.ai outperformed the other models by achieving the best overall balance between detecting harmful content and avoiding false positives — a critical trait for real-world moderation applications. Claude showed high toxic recall but poor non-toxic precision, leading to over-flagging, while OpenAI and Mistral underperformed in handling culturally nuanced toxicity, highlighting the advantage of Typica.ai's culturally tailored approach.

## 5.1 Qualitative Error Analysis

Qualitative analysis of the model outputs revealed that OpenAI, Mistral, and Claude frequently failed to flag harmful content embedded in localized idioms, cultural sarcasm, or implicitly aggressive tones common in Moroccan Darija.

For example, the following user-generated comment was consistently misclassified as benign by OpenAI and Mistral, while Claude flagged it as toxic but also produced many unrelated false positives:

> "هدا مافنان ماهو راجل ماهو مرا باختصار الله ينعل الأصل ديال الي رباه"
> *(roughly, "This one is neither a man nor a woman, simply put: may God curse whoever raised him" — a culturally charged insult)*

These types of expressions, although not explicitly profane, carry strong cultural and contextual toxicity, making them difficult for general-purpose models to detect. Overall, OpenAI, Mistral, and Claude performed adequately on overt and globally recognizable toxic patterns but struggled with culturally nuanced or indirect toxicity. In contrast, Typica.ai's model, trained specifically on Moroccan Darija, successfully captured these indirect and contextually encoded insults. While Typica.ai's classifier did produce some false positives, they generally involved highly sarcastic but ultimately benign messages, highlighting an opportunity for future sarcasm-aware calibration.

## 5.2 Comparative Strengths and Weaknesses

We conclude the Results section by summarizing key strengths and weaknesses across the evaluated moderation models. Typica.ai achieved the highest overall accuracy (83%) and the best balance between precision (0.856) and recall (0.812) on the toxic class, demonstrating its strength in reliably detecting harmful content without excessive false positives — a critical trait for real-world moderation.

In contrast, OpenAI and Mistral showed comparable performance, with reasonably high recall (~0.84) on non-toxic content but notably low recall (~0.47) on toxic content, indicating they frequently missed culturally nuanced or indirect toxicity. Claude, on the other hand, exhibited very high toxic recall (0.942) but low non-toxic precision, leading to over-flagging and misclassifying many benign comments as harmful.

These findings highlight an important trade-off in content moderation: specialized models like Typica.ai can achieve robust performance in low-resource cultural contexts through domain-aware training, while general-purpose APIs may struggle without local adaptation. Future improvements could focus on fine-tuning general models for culturally specific patterns or further calibrating specialized models like Typica.ai to reduce false positives, particularly on benign sarcasm or nuanced expressions.

## 6. DISCUSSION

The outcomes of this study highlight substantial limitations in general-purpose LLM-based moderation models when applied to dialect-rich languages such as Moroccan Darija. To the best of our knowledge, Typica.ai's culturally targeted model demonstrates significant improvements in moderation accuracy, underscoring the critical importance of localization and domain-specific adaptation. While previous studies have primarily evaluated models in isolation, without direct benchmarking against widely used commercial LLM-based moderation APIs, our comparative approach helps bridge this gap by providing empirical evidence that localized models can outperform broad multilingual systems on culturally nuanced tasks.

Notably, this benchmark sheds new light on the trade-offs between broad multilingual generalization and culturally specific sensitivity, reinforcing the argument that one-size-fits-all LLM moderation approaches often miss critical local subtleties. Importantly, this discussion also reflects on the contrasts between smaller language models (SLMs), such as our BERT-based Typica.ai classifier, and large LLM-based moderation systems.

While LLMs like OpenAI, Mistral, and Anthropic Claude provide broad language coverage, they often sacrifice local nuance. Claude's performance, in particular, illustrates both the potential of prompt-driven moderation — where toxic recall can be high — and the challenge of calibration, as seen in its elevated false positive rates.

Additionally, future research should investigate how LLM-based systems can be complemented by smaller, domain-adapted models to better capture fine-grained, culturally specific tones — such as sarcasm, mockery, and social innuendo — that are particularly sensitive in Moroccan society. Beyond hybrid architectures, future work should also assess the cost-performance trade-offs between large-scale LLM deployments and lightweight, culturally specialized models, particularly for resource-constrained or real-time applications. Moreover, integrating human-in-the-loop pipelines could enhance cultural sensitivity and ethical oversight in automated moderation workflows. Finally, continuous data curation and model updating will be essential to ensure these systems remain aligned with evolving language patterns and community norms over time.

## 7. LIMITATIONS

While this study provides meaningful insights, several limitations must be acknowledged to contextualize the findings. First, although the OMCD dataset and additional sources were carefully curated, they contain examples not strictly Moroccan but drawn from other Arabic dialects (such as Middle Eastern variants), potentially introducing cross-dialectal noise that may affect evaluation outcomes. Second, the annotation process, like many in the field, is inherently subjective: definitions of toxicity are culturally shaped and can vary among annotators, which may introduce bias or inconsistencies.

To strengthen transparency and adhere to reproducibility best practices, we have made the evaluation notebook, test split, and model prediction outputs publicly available through a dedicated GitHub repository [7], enabling independent benchmarking against the reported results. While internal model weights and proprietary training data are not released, we plan to offer controlled API access — following the approach of providers like OpenAI, Mistral, and Anthropic Claude — to allow third-party researchers and practitioners to reproduce the evaluation pipeline and conduct extended assessments.

## 8. CONCLUSION

This study presents a comprehensive comparative benchmark focused on Moroccan Darija toxicity detection, offering one of the first systematic evaluations of culturally adapted models against widely adopted commercial moderation tools. Our custom-built Typica.ai classifier consistently outperformed leading moderation APIs from OpenAI, Mistral, and Anthropic Claude, providing robust empirical evidence that culturally specialized models can surpass general-purpose LLM-based systems in dialect- and culture-specific tasks.

These results reinforce findings from prior research highlighting that large multilingual LLMs, while powerful, often struggle to generalize effectively to dialectal and culturally embedded content without explicit domain adaptation. They also contribute to the growing recognition that culturally grounded benchmarks are essential for advancing fair and inclusive NLP systems, particularly in historically underrepresented regions.

Importantly, beyond performance improvements, these findings carry ethical implications: they highlight the need for moderation systems that respect cultural nuance, reduce bias, and ensure fair treatment for diverse online communities. As demonstrated by the Typica.ai classifier, domain-specific models offer a promising, cost-effective pathway to addressing the limitations of broad multilingual systems in real-world moderation scenarios, particularly for low-resource languages like Moroccan Darija.

## 9. ACKNOWLEDGMENTS

This work at Typica.ai builds on contributions to Moroccan Darija toxicity datasets, including OMCD. We also thank OpenAI and Mistral for providing free-access moderation APIs that enabled fair benchmarking.